\documentclass[11pt]{article}

\usepackage[final]{acl}

\usepackage{times}
\usepackage{latexsym}
\usepackage{booktabs}

\usepackage[T1]{fontenc}

\usepackage[utf8]{inputenc}

\usepackage{microtype}

\usepackage{inconsolata}

\usepackage{graphicx}
\usepackage{xspace}

\newcommand{\benchmark}{\textsc{PeerMathDial}\xspace}

\title{\benchmark: A Middle School Dialogue Dataset for Student Collaborative Math Problem Solving}

\author{
 \textbf{Murong Yue\textsuperscript{1}},
 \textbf{Desmond Alexander Mcglone\textsuperscript{2}},
 \textbf{Emily Slutz\textsuperscript{2}},
 \textbf{Wenhan Lyu\textsuperscript{3}},
\\
 \textbf{Yixuan Zhang\textsuperscript{3}},
 \textbf{Jennifer Suh\textsuperscript{2}},
 \textbf{Ziyu Yao\textsuperscript{1}}
\\
\\
 \textsuperscript{1}\normalsize Department of Computer Science, George Mason University\\
 \textsuperscript{2}\normalsize College of Education \& Human Development, George Mason University\\
 \textsuperscript{3}\normalsize Department of Computer Science, William \& Mary\\
 \footnotesize \textbf{Project website (dataset and source code):} \href{https://ziyu-yao-nlp-lab.github.io/MathVC-NSF.github.io/}{MathVC-NSF.github.io}
\\
}

\begin{document}

\maketitle
\begin{abstract}
Collaborative Problem Solving (CPS) is a core skill in education, where the process of peer interaction is highly important. However, existing educational dialogue datasets mostly focus on classroom instruction or tutoring (i.e., teacher/tutor-student interaction), yet datasets centering small-group, student-student interaction are limited. This thus leaves research with limited resources for studying how students interact, coordinate, and solve problems together in real educational settings. 
To address this, we introduce \benchmark, the first dataset of peer CPS dialogues collected from authentic middle school math classrooms. It contains 55 dialogues from 27 students, totaling 6,406 turns.
To facilitate research on CPS discourse analysis, we further build a corpus-grounded dialogue act taxonomy assisted by LLMs.
Using the dataset and the dialogue act taxonomy, we demonstrate the practical applications of \benchmark across three use cases. First, we track how dialogues evolve over time and measure the impact of teacher interventions. Second, we align dialogue actions with student surveys to reveal the connection between students' traits (e.g., confidence, leadership) and their actual behaviors. Third, by evaluating LLMs on dialogue act prediction, we glimpse at the potential of LLMs for student simulation in educational applications.
Our dataset and source code will be released to the community.

\end{abstract}

\section{Introduction}

Collaborative Problem Solving (CPS) is recognized as a core competency in K-12 education, as students share information, coordinate participation, explain reasoning, check constraints, and jointly revise errors through interaction~\citep{vygotsky1978mind,rose2008analyzing,care2012assessment}. 
In contrast to product‑oriented problem‑solving processes, an exclusive focus on outcomes fails to capture how students actually collaborate through dialogue, including how they negotiate meaning, consider different perspectives, generate solutions, and respond to one another in interaction.
Although the importance of CPS as a process‑oriented approach is well recognized in the mathematics literature~\citep{roschelle1995construction, o2013introduction, pisa2018assessment}, access to authentic classroom dialogue that enables fine‑grained examination of these processes remains limited.
Important educational dialogue resources such as TalkMoves~\citep{suresh2022talkmoves}, NCTE~\citep{demszky-hill-2023-ncte}, CIMA~\citep{stasaski2020cima}, and MathDial~\citep{macina2023mathdial} are respectively more oriented toward classroom instructional discourse, tutoring dialogue, and constructed educational interaction, yet they do not provide dialogues centered on peer interactions in CPS. 

To address this gap, we introduce \benchmark, a CPS dialogue dataset collected from a math-centered, middle-school summer camp. The dataset captures students' naturally occurring face-to-face small-group problem solving and contains 55 unscripted dialogues from 27 students, comprising 6,406 dialogue turns in total. 
Our dataset preserves the spoken interactional process of classroom collaboration, including task interpretation, strategy proposal, peer questioning, etc. 
To the best of our knowledge, we are the first to publish a dataset of face-to-face peer CPS dialogues collected from authentic K-12 mathematics classrooms.
Besides, to make the corpus more analyzable, we further derive a dialogue act~\citep{stolcke2000dialogue} taxonomy to organize students' interactions. 
Rather than mapping dialogue turns onto expert-designed taxonomies such as the PISA CPS framework~\citep{pisa2018assessment}, we adopt an LLM-assisted, corpus-grounded inductive procedure: an LLM first processes the raw dialogue and summarizes dialogue actions, after which these candidate actions are manually reviewed by education experts. 
The resulting taxonomy covers major interactional dimensions such as affective/off-task talk, communicating and coordinating, exploring and understanding, monitoring and revising, planning and executing, and representing and formulating. 
Its purpose is to provide a corpus-grounded organizational lens that makes subsequent educational dialogue analysis more tractable.

Beyond introducing the corpus and its action taxonomy, we demonstrate the applications of \benchmark through three specific applications. 
(1) \textbf{Understanding collaborative dynamics}: we track the temporal evolution of dialogues and empirically quantify the impact of teacher interventions on students' subsequent micro-behaviors. 
(2) \textbf{Understanding the connection of actions and characters}: by aligning dialogue act distributions with self-reported character surveys (including math confidence, collaboration, leadership), we reveal typical patterns between students' character and their behaviors in CPS. 
(3) \textbf{Evaluating LLM student simulation}: through dialogue act prediction experiments, we show that state-of-the-art LLMs still struggle to reproduce authentic collaborative behaviors, establishing our dataset as an ideal testbed for developing high-fidelity virtual student simulators.

\section{Dataset}
\subsection{Data Collection}
Our dataset was collected from a summer camp organized by George Mason University in 2025, centered on middle-school (i.e., grade 6-8) mathematical CPS. 
During each session, students participated in a 90-minute collaborative problem-solving block, in which they worked in small teams of 3-4 students on two to three mathematical tasks. To guide the session, the camp recruited 4 middle-school teachers from nearby school districts with an average of 20 years of teaching experience. The mathematical tasks were designed by these teachers, balancing between difficulty and engagement.
Participants enrolled in the camp through recruitment materials distributed via the university's camp website, social media platforms, and educational mailing lists.
More details about the participants' information and the camp tasks are in the Appendix~\ref{appendix: student demographic}.

The instructional design emphasized open-ended, real-world-inspired mathematical tasks intended to elicit collaborative reasoning, mathematical modeling, and collective sense-making. Across the week, students engaged in tasks that required them to propose solution strategies, justify their reasoning, and revise their ideas in response to peer feedback. For example, one task asked students to determine how to price a cake fairly based on its size under a fixed total price, requiring proportional and spatial reasoning. Together, these tasks were designed to foreground discussion, explanation, conjecturing, and revision rather than answer production alone.

All instructional sessions were video- and audio-recorded to capture both whole-class instruction and small-group interaction. We used Swivl cameras for whole-class video capture and placed an individual voice recorder at each table to obtain higher-quality audio of small-group collaborative dialogue. These recordings form the basis of the benchmark.

\subsection{Data Processing and Statistics}
To prepare the corpus, we maintained a centralized data log that tracked all available recordings by date, classroom, task, and group. 
We performed speech recognition using Otter AI.\footnote{\url{https://otter.ai/}} All recordings were first transcribed automatically and then manually reviewed by four student researchers while listening to the original audio. We followed a two-level transcription workflow: (1) whole-lesson transcription, which captured teacher-led and class-wide discourse, and (2) small-group transcription, organized by table or group identity to preserve fine-grained peer interaction. Because Otter AI provides time-aligned transcripts, researchers were able to revise transcripts while checking the original recordings. We made light but systematic edits to improve lexical accuracy, punctuation, and readability, while preserving the original turn-by-turn structure of the interaction.
Speaker identification was conducted manually. Otter AI initially assigned generic labels such as \textit{Speaker 1} and \textit{Speaker 2}; researchers replaced these labels with the corresponding student or teacher identities using session record sheets, group assignments, seating information, and audio/video context. During transcript cleanup, utterance segments that had been incorrectly split by the automatic system were merged to maintain speaker coherence and conversational flow. Real names were retained during the internal transcription stage to support cross-session tracking, and all transcripts were pseudonymized in later analytic stages.
This processing pipeline was designed to preserve the temporal and interactional structure of collaborative dialogue, enabling analysis of participation, reasoning, and coordination in small-group mathematical problem solving.

\begin{table}[t!]
\centering
\small
\begin{tabular}{lc}
\toprule
\textbf{Metric} & \textbf{Number} \\
\midrule
Total Student & 27 \\
Total Dialogues & 55 \\
Total Dialogue Turns & 6,406 \\
Average Turns per Dialogue & 116.5 \\
Average Words per Turn & 11.6 \\
\midrule
\multicolumn{2}{l}{\textbf{Role Distribution}} \\
\quad Student & 71.7\% \\
\quad Teacher & 20.7\% \\
\quad Unknown & 7.6\% \\
\bottomrule
\end{tabular}
\caption{Statistics of the Dialogue Dataset.}
\label{tab:dataset_statistics}
\end{table}

The raw classroom transcripts contained overlapping whole-class instructional content. To only extract authentic peer interaction within small groups, we remove shared whole-class segments that were duplicated across multiple group transcripts, such as teacher-led openings at the beginning of class or summary discussions at the end. This process ensured that the final retained data consisted only of localized CPS dialogue within each group. In addition, we conducted a rigorous anonymization procedure, mapping original speaker names to consistent pseudonyms across all sessions. This allowed us to accurately track individual participation longitudinally while preserving privacy.

The statistics of the cleaned dataset are summarized in Table~\ref{tab:dataset_statistics}. The dataset contains 55 unique dialogue sessions with a total of 6,406 conversational turns. On average, each session includes 116.5 turns, and each turn contains 11.6 words, reflecting the relatively fast-paced and fragmented nature of middle school conversational exchanges. In terms of participation, the dataset tracks 27 focal students who completed self-report background surveys. Analysis of the role distribution indicates a student-driven CPS environment, with student utterances accounting for the vast majority of the discourse (71.7\%). Finally, 7.6\% of the turns are labeled as ``unknown,'' typically due to brief interjections, overlapping speech, or background noise that could not be reliably attributed to a specific speaker.

\section{CPS Dialogue Act Discovery}
\begin{table*}[t]
\centering
\small
\begin{tabular}{p{3.8cm}p{11cm}}
\toprule
\textbf{Stage} & \textbf{Dialogue acts} \\
\midrule
Affect and Off-Task Talk 
& Off-Task or Affective Commentary \\
\midrule
Communicating and Coordinating 
& Acknowledging or Receiving; Explaining or Justifying Reasoning; Guiding Peer Participation or Shared Workspace Actions; Presenting Group Reasoning to an External Listener \\
\midrule
Exploring and Understanding 
& Asking for Clarification or Explanation; Identifying Relevant Elements and Constraints; Orienting to Lesson or Activity \\
\midrule
Monitoring and Revising 
& Challenging or Diagnosing an Error; Checking What a Number or Step Represents; Checking Whether a Result Fits Constraints; Revising the Method or Answer \\
\midrule
Planning and Executing 
& Carrying Out a Computation Step; Decomposing or Repartitioning the Structure; Proposing a Candidate Solution; Proposing a Solution Strategy \\
\midrule
Representing and Formulating 
& Constructing or Refining a Representation; Mapping Representation to Values; Proposing a Pattern or Structural Model; Reusing a Previous Case or Pattern \\
\bottomrule
\end{tabular}
\caption{Overview of the corpus-grounded CPS dialogue act taxonomy. Full dialogue act definitions and examples are provided in Appendix~\ref{appendix: action taxonomy}.}
\label{tab:dialogue_acts}
\end{table*}
Dialogue acts are functional units describing the speaker's intent \citep{stolcke2000dialogue}, which have been widely used for discourse analysis and conversational system construction.
Educational research on CPS discourse has often relied on pre-defined coding frameworks, such as the PISA framework~\citep{pisa2018assessment, stadler2020assessment}. These expert-developed frameworks typically focus on relatively coarse-grained behavioral categories, such as ``establishing and maintaining shared understanding.'' However, it cannot be directly applied to authentic middle school discussions. Student utterances are typically short and colloquial, and the granularity of the PISA framework is often too coarse for turn-level analysis. For example, questioning a problem condition and discussing a solution strategy may both be mapped to ``establishing and maintaining shared understanding'' under the PISA framework, despite serving different interactional dialogue act in local classroom discourse. To address this limitation, we adopt a bottom-up, LLM-assisted dialogue act induction pipeline that summarizes and organizes recurring interactional behaviors directly from the corpus.
Table~\ref{tab:dialogue_acts} summarizes the resulting taxonomy with details in Appendix~\ref{appendix: action taxonomy}.

This pipeline is to leverage the LLM to identify the dialogue acts of each student turn within its full conversational environment. It consists of three steps. First, we divide the corpus into batches, each containing multiple dialogues, and submit the batch one by one to the LLM. For each batch, the model is asked to summarize the dialogue acts presented in this batch. Through this induction process, we extract repeated act descriptions across dialogues and gradually form a set of candidate dialogue acts. We then consolidate these batch-level summarized dialogue acts and explicitly control the granularity through prompting, so as to avoid acts that are overly broad (e.g., ``establishing shared understanding'') or overly specific (e.g., ``saying ok''), resulting in an initial act inventory tailored to this corpus.
Second, after obtaining the initial act inventory, we ask the LLM to assign a dialogue act to every student turn in the dataset. For each turn, the model is required not only to provide an act label but also to produce a verbal confidence uncertainty score indicating how certain it is about the assignment~\citep{he2025survey}. This allows us to identify the cases that the model itself considers uncertain.
Third, we collect the turns with low confidence from the first round and submit them back to the LLM. At this stage, our goal is to examine whether these high-uncertainty cases reveal additional dialogue acts that are not yet covered by the initial act inventory. Through this focused inspection of low-confidence turns, the LLM can summarize potential new acts and thereby extend or revise the taxonomy.
After obtaining the updated act inventory, we use the LLM to perform a new round of turn-level labeling over the full corpus. Overall, this pipeline combines corpus-level act discovery, low-confidence-case-driven taxonomy refinement, and full-corpus relabeling, thereby reducing the risk of missing dialogue act types as much as possible. All LLMs used are GPT-5.4~\citep{openai_gpt54_2026}.

To evaluate the quality of the induced taxonomy and its turn-level labeling, we conducted an expert evaluation with two math education researchers who have multi-year experience as school teachers. The evaluation included two parts. In part 1, the experts were presented with the taxonomy definitions together with multiple examples and were asked if they agreed on them. In part 2, the experts evaluated 100 conversational turns, comprising 50 randomly sampled instances and 50 instances marked by the LLM as low-confidence. For each turn, we also provided the surrounding conversational context to disambiguate the judgment. The experts were then asked to judge whether the assigned act adequately described the student’s turn; if not, they were further asked whether a better label could be found in the taxonomy, or if the taxonomy was missing dialogue acts to cover the turn. As such, the experts were positioned to both \emph{review the LLM assignment} and \emph{make independent annotations}.
Our results indicate that the dialogue act taxonomy provides strong coverage of the interactional behaviors in the corpus. The experts agreed with the LLM's assignments for all 50 randomly sampled turns and only identified 11 misclassifications of existing acts (rather than missing acts in the taxonomy) within the low-confidence set.
Overall, the expert evaluation supports the validity of the induced taxonomy and suggests that this LLM-assisted pipeline is able to capture the major CPS interactional behaviors in the corpus with substantial coverage.
We include details of the expert evaluation in Appendix~\ref{appendix: expert review}.

\section{Application I: Enhancing Educational Understanding of the CPS Process}
\begin{figure*}[t]
    \centering
\includegraphics[width=0.95\textwidth]{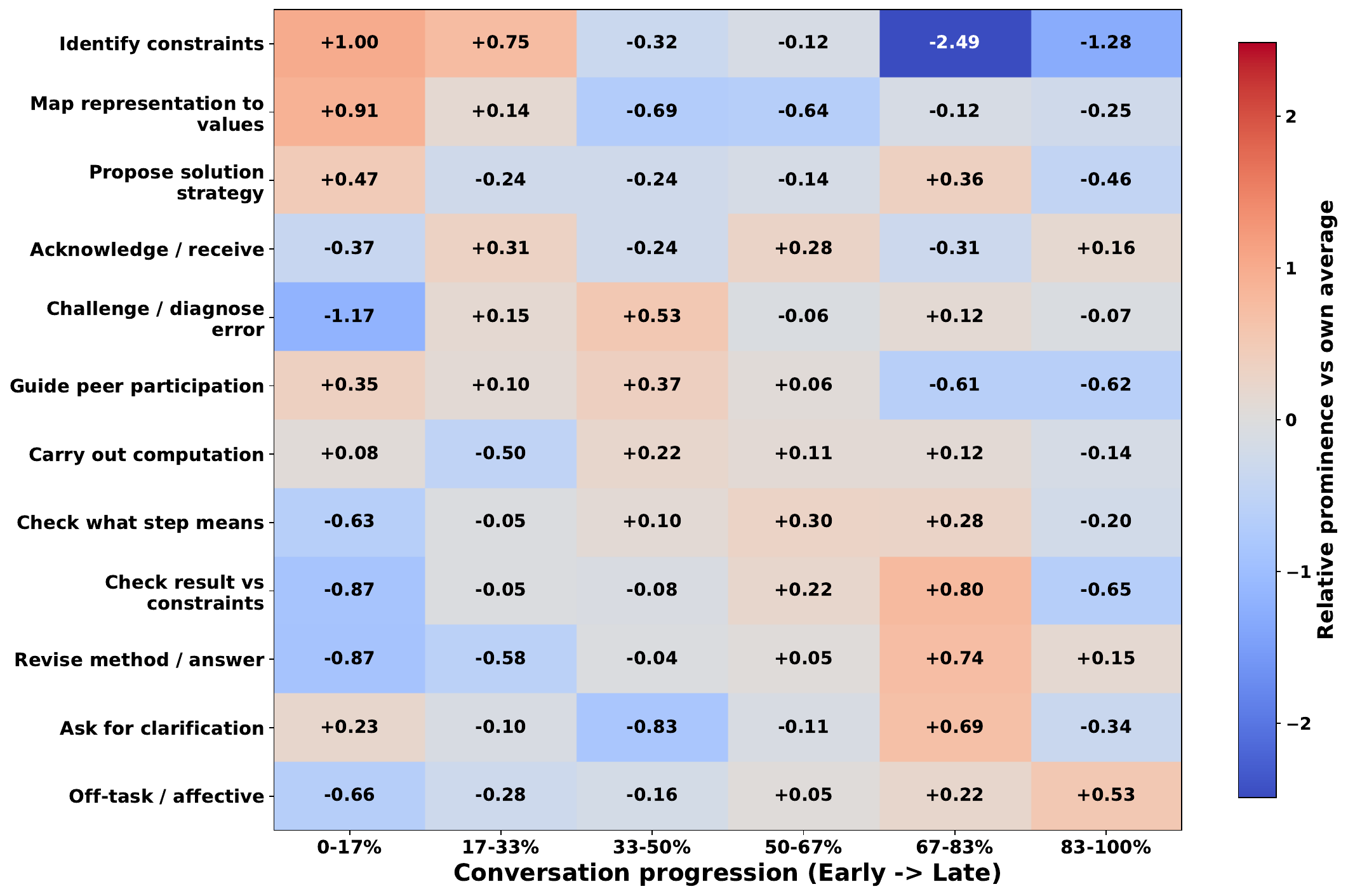}
\caption{Temporal evolution of student dialogue acts over the course of CPS conversations. 
Each dialogue is uniformly divided into six stages based on relative progress. Let $p_{a,s}$ denote the proportion of act $a$ within stage $s$, and $\bar{p}_a$ represent its overall average proportion across all stages. The heatmap visualizes the log-ratio, $\log_2(p_{a,s} / \bar{p}_a)$.
Positive values indicate above-typical prominence, and negative values indicate below-typical prominence.}
    \label{fig:temporal-evolution-dialogue-acts}
\end{figure*}
The discovered dialogue acts offer a lens through which educational researchers can better understand CPS. We present two possible directions in this section.
\subsection{Temporal Evolution of CPS Dialogue Acts}
We first investigate how the frequency of specific dialogue acts changes as a conversation unfolds. To compare dialogues of different lengths, we divide each conversation into six equal segments based on relative progress and compute dialogue-act proportions using student turns only. For each act, we first compute its proportion within each segment of each conversation and then average these proportions across conversations, preventing longer dialogues from dominating the estimate. For visualization, we express each segment's value as a log-scaled deviation from that act's average level across the full conversation. Under this normalization, positive values indicate greater-than-usual prominence, and negative values indicate lower-than-usual prominence.

Figure~\ref{fig:temporal-evolution-dialogue-acts} reveals a clear temporal structure in the CPS process when we align student turns by relative dialogue progress. 
Overall, the temporal pattern suggests that CPS dialogue is not static: it moves from early problem framing and coordination, to mid-stage calculation and error diagnosis, and finally to late-stage evaluation, revision, and convergence.
Early stages are characterized by actions that help establish the problem space, such as \textit{identifying relevant elements and constraints}, \textit{mapping representation to values}, and \textit{proposing a solution strategy}. 
At the same time, \textit{guiding peer participation or shared workspace actions} is also more prominent near the beginning, suggesting that students initially devote effort to coordinating participation and setting up a shared working process.
As the interaction moves toward the middle stages, the dialogue becomes more operational and diagnostic. 
Actions such as \textit{challenging or diagnosing an error}, \textit{carrying out a computation step}, and \textit{checking what a number or step represents} become relatively more prominent, indicating a shift from task setup toward active solution development and local monitoring of intermediate reasoning. 
This middle portion appears to function as the main problem-solving core, where students test ideas, compute, and interrogate the meaning of emerging results. 
In later stages, the discourse increasingly shifts toward evaluation and repair. 
\textit{Checking whether a result fits constraints} and \textit{revising the method or answer} become especially salient, consistent with a closing phase in which students verify whether their current solution is acceptable and make adjustments when necessary.
Late-stage increases in \textit{asking for clarification or explanation} suggest that final convergence often still requires resolving residual uncertainty rather than simply stating a completed answer. 
We also observe a modest rise in \textit{off-task or affective commentary} toward the end, which may reflect easing task pressure once the main solution path has been established.

\subsection{Impact of Teacher Intervention}
\begin{figure*}[t]
\centering
\includegraphics[width=1\linewidth]{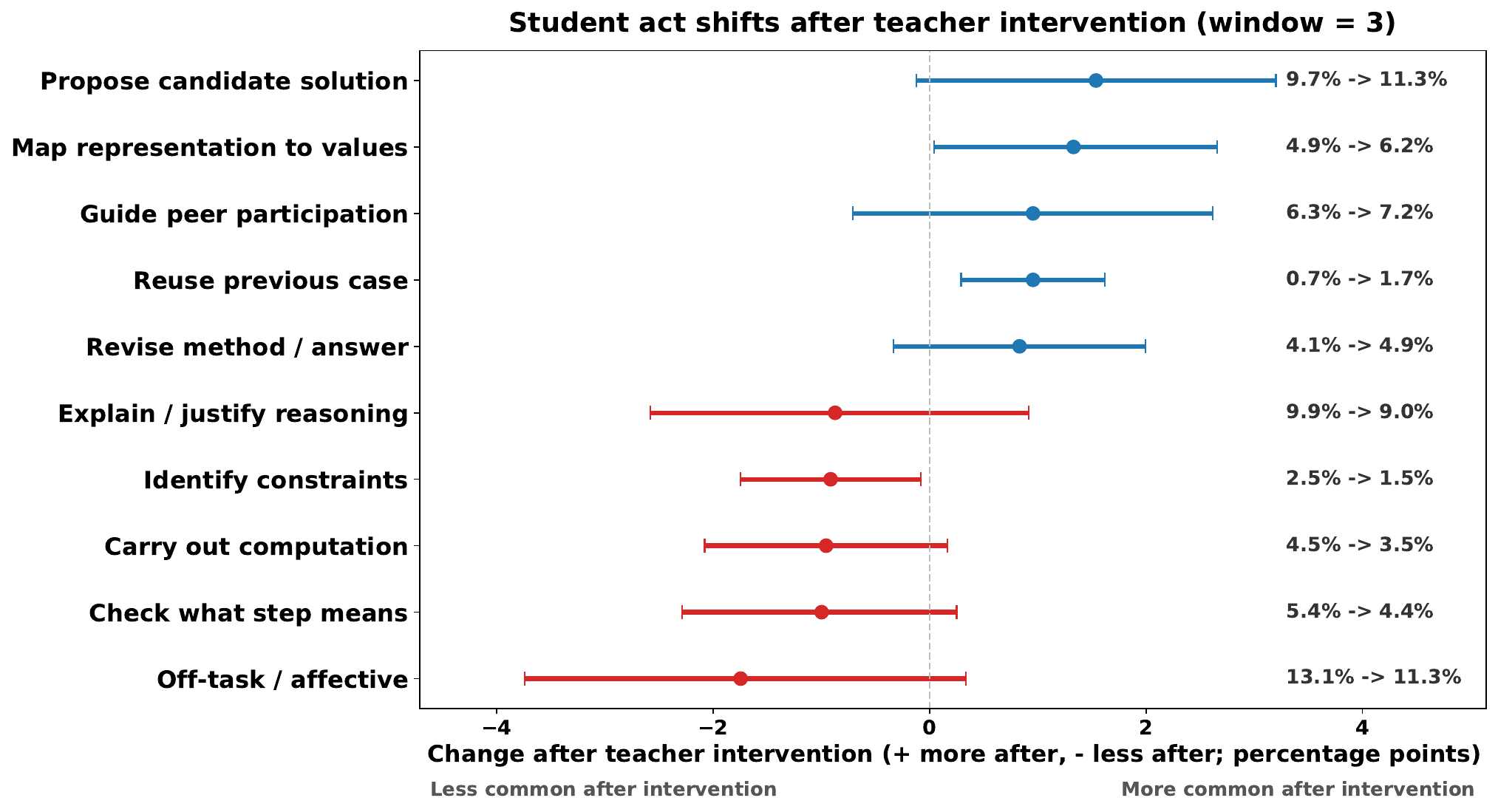
}
\caption{Change in student dialogue-act prevalence before versus after teacher intervention, using a three-turn student window on each side of the intervention.
}
\label{fig:teacher_intervention_shift}
\end{figure*}

Teacher intervention is an important component of CPS in educational settings, as teachers can redirect attention, clarify task demands, prompt reflection, and help groups move past local impasses. To better understand this role, we examine how students' dialogue acts shift immediately before and after teacher intervention. We operationalize each contiguous run of teacher turns as a single intervention event, and for each event, extract the three nearest annotated student turns immediately before it and three immediately after it. For every intervention event and every dialogue act, we compute the act's share in the pre-window and in the post-window, and define the event-level shift as \emph{post share minus pre share}.

Figure~\ref{fig:teacher_intervention_shift} reports the mean of this post-minus-pre quantity across intervention, measured in percentage points: values to the right of zero indicate acts that become more common after teacher intervention, whereas values to the left indicate acts that become less common. Horizontal lines show 95\% bootstrap confidence intervals. This setup allows us to characterize teacher intervention not as a single isolated response, but as a short-run redistribution of student activity in the turns that follow.
The post-intervention shows reliable increases in \textit{mapping representation to values} (from 4.9\% to 6.2\%) and \textit{reusing a previous case or pattern} (from 0.7\% to 1.7\%), together with a reliable decrease in \textit{identifying relevant elements and constraints} (from 2.5\% to 1.5\%). Descriptively, we also observe a shift toward \textit{proposing a candidate solution} and away from \textit{off-task or affective commentary}, although those trends are less certain because their confidence intervals overlap zero. The core takeaway is that teacher facilitation appears to redirect the next few student turns away from local interpretation and toward forward-moving mathematical activity.

\section{Application II: Facilitating Teacher Comprehension of Student Profiles}
\begin{figure*}[t]
\centering
\includegraphics[width=1\linewidth]{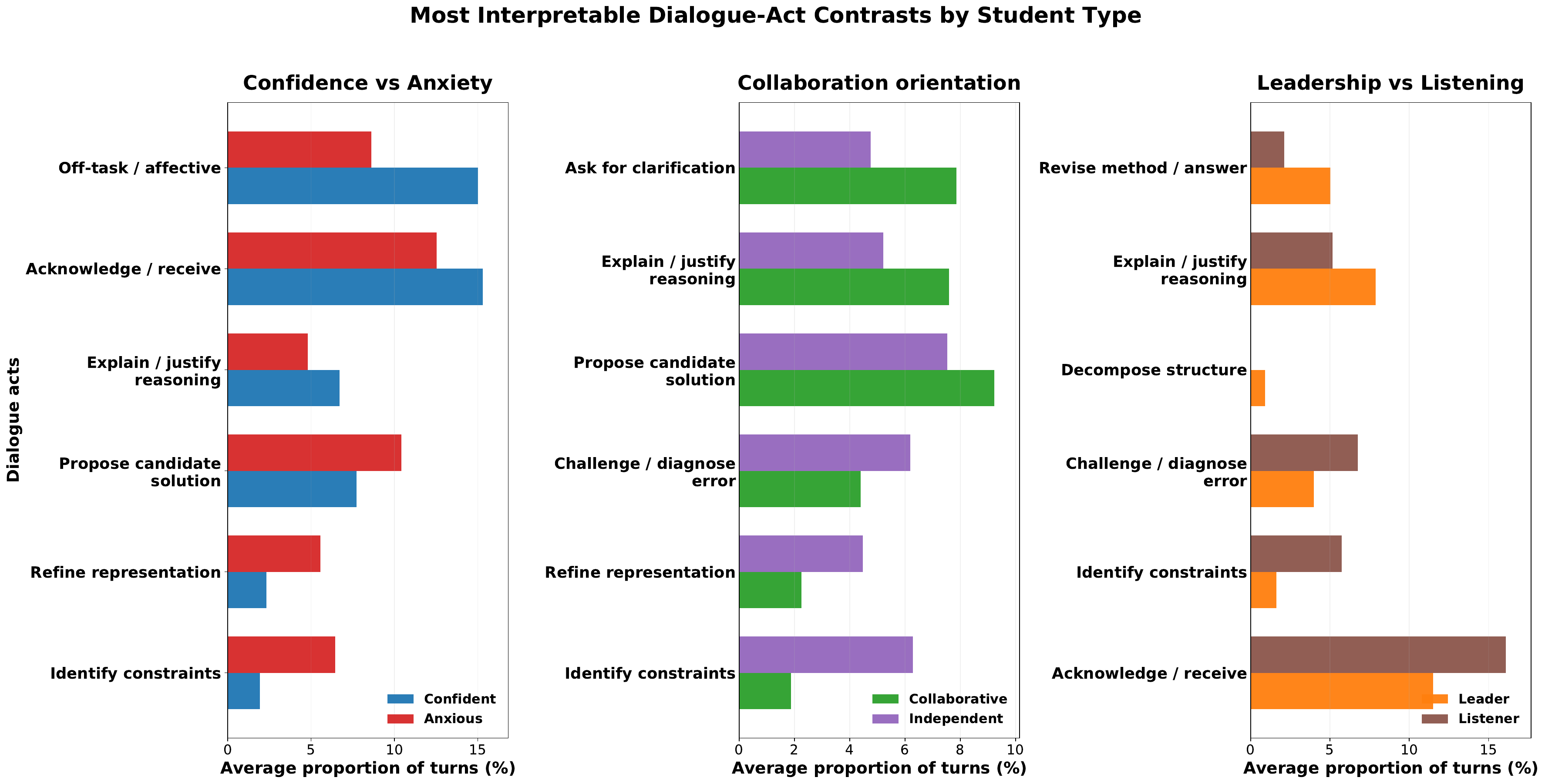}
\caption{Comparison of dialogue-act usage. Students are grouped from pre-task survey responses along three dimensions: confidence (Confident vs.\ Anxious), collaboration orientation (Collaborative vs.\ Independent), and participation stance (Leader vs.\ Listener). Bar length shows the average proportion of each student's annotated turns assigned to a given dialogue act.
}
\label{fig:student_typology_action_heatmap}
\end{figure*}

Beyond the understanding of the conversational process, dialogue acts serve as a powerful tool for understanding individual student characteristics. 
By aligning the dialogue acts distribution with self-reported surveys, we provide educators with a multi-dimensional view of student engagement.
Student typologies are derived from a pre-program self-report survey. The survey consists of Likert-scale items with five response options ranging from \textit{strongly disagree} to \textit{strongly agree}, covering students' attitudes toward mathematics, collaboration, leadership, and related dispositions. For each profile dimension used in our analysis (e.g., confidence, collaboration, leadership), we group together the relevant survey items and assign students to one of two broad types based on whether their responses overall indicate endorsement of that dimension. In other words, students with more agreement-oriented responses on a given dimension are grouped into the corresponding type (e.g., \textit{confident}, \textit{collaborative}, or \textit{leader-like}), whereas students without such endorsement are grouped into the contrasting type. Full survey items and their grouping are provided in Appendix~\ref{appendix :survey_profiles}.

\paragraph{Dialogue Act Distributions Across Student Typologies}
Figure~\ref{fig:student_typology_action_heatmap} shows the dialogue act distribution by student types.
The results suggest that the survey dimensions capture differences in \emph{how} students participate, rather than a single continuum of general participation. The clearest contrast appears along the leadership dimension. Compared with listeners, leader-type students produce more \textit{revising the method or answer} (5.0\% vs.\ 2.1\%) and more \textit{explaining or justifying reasoning} (7.9\% vs.\ 5.2\%). By contrast, listener-type students contribute more \textit{acknowledging or receiving} (16.1\% vs.\ 11.5\%, difference favoring listeners by $4.58$ percentage points) and more \textit{identifying relevant elements and constraints} (5.7\% vs.\ 1.6\%). A similar functional differentiation appears along the collaboration dimension: collaborative students more often engage in \textit{asking for clarification or explanation} (7.9\% vs.\ 4.8\%) and \textit{explaining or justifying reasoning} (7.6\% vs.\ 5.2\%), whereas independent students more often foreground \textit{identifying relevant elements and constraints} (6.3\% vs.\ 1.9\%). Along the confidence dimension, confident students show a higher rate of \textit{off-task or affective commentary} (15.0\% vs.\ 8.6\%), while anxious students more often produce \textit{identifying relevant elements and constraints} (6.5\% vs.\ 1.9\%, difference favoring anxious students by $4.54$ percentage points) and \textit{constructing or refining a representation} (5.6\% vs.\ 2.3\%, difference favoring anxious students by $3.25$ percentage points). Taken together, these patterns indicate that quieter or less leader-like students are not simply disengaged; rather, they tend to contribute through uptake, monitoring, and local grounding moves, whereas leader-like and collaborative students contribute disproportionately through explanation, clarification, and revision.

\paragraph{Discrepancies in Self-Reported Surveys}
At the same time, the dialogue-act analysis highlights an important limitation of treating self-reported survey responses as direct proxies for behavior. One particularly revealing case is the item \textit{``I try to avoid arguments, even if I disagree.''} If interpreted literally, stronger agreement with this item might be expected to predict lower rates of \textit{challenging or diagnosing an error}. However, the observed discourse does not follow that simple pattern. Among students with sufficient dialogue data, those who \textit{agreed} or \textit{strongly agreed} with the argument-avoidance item still produced \textit{challenging or diagnosing an error} in 5.9\% of their annotated turns, compared with 3.2\% for students who gave a \textit{neutral} response. One possible interpretation is that students understand ``avoiding arguments'' as avoiding interpersonal conflict rather than refraining from mathematical critique. Another is that such behaviors are shaped by context: in relatively low-stakes tasks, students may perceive less social risk in pointing out an error. More broadly, this discrepancy underscores that survey responses capture self-perception and stated disposition, whereas dialogue acts capture situated behavior in collaborative activity. The two are related, but not isomorphic. For this reason, combining survey-based typologies with discourse-analytic evidence provides a more reliable account of student participation than either source alone.

\section{Application III: Implications for LLM Student Simulation}
\begin{table}[t]
\centering
\small
\caption{Agreement between LLM-predicted actions and real student actions under naive student-type prompting, evaluated on 100 sampled cases.}
\label{tab:llm-student-simulation-results}
\begin{tabular}{lc}
\toprule
Model & Match Rate \\
\midrule
OpenAI GPT-5.4 Mini & 16.0\% \\
OpenAI GPT-5.4 & 18.0\% \\
Anthropic Claude Sonnet 4.6 & 16.0\% \\
Google Gemini 3 Flash Preview & 20.0\% \\
Qwen3.5-35B-A3B & 17.0\% \\
\bottomrule
\end{tabular}
\end{table}

In the field of educational AI, using LLM to simulate real students for interaction holds significant value~\citep{yue2024mathvc}. However, constructing high-fidelity student simulators remains highly challenging~\citep{kumar2025can}. Virtual students must not only produce natural responses in dialogue, but their behavioral decisions must also align with the real students. \benchmark provides a highly valuable ground truth for developing and evaluating such simulators.

To conduct an investigation of current LLMs' ability, we designed an action prediction experiment. We randomly sampled 100 dialogue turns and provided the LLM with only the student survey and all utterances of the dialogue context. Without using complex scaffolding techniques, we asked the model to predict the most likely next action from the predefined dialogue act taxonomy. We evaluated five representative state-of-the-art models, including GPT-5.4, GPT-5.4-mini~\citep{openai_gpt54_2026}, Claude Sonnet 4.6~\citep{anthropic_claude_sonnet46_2026}, Gemini 3 Flash Preview~\citep{google_gemini3_flash_2026}, and the open-source Qwen3.5-35B-A3B~\citep{qwen35_35b_a3b_modelcard_2026}.

The experimental results in Table~\ref{tab:llm-student-simulation-results} indicate that current LLMs struggle to accurately reproduce the fine-grained collaborative behaviors of real students when provided with only basic identity prompts and short context windows. Across all test cases, the exact match rate of predicted actions for each model ranged only between 16.0\% and 20.0\%. 
While we admit that in some contexts more than one dialogue act can be possible, this very low exact match rate still implies a mismatch between the LLM prediction and the majority of the dialogue development.
This finding carries important implications for future educational applications of LLMs. If we aim to build virtual students with behaviorally realistic decision-making processes, relying solely on high-level persona attributes is insufficient. Future student simulators may require deeper fine-tuning, longer-term dialogue memory, or specialized modeling of mathematical collaboration states. The dataset we provide serves as an ideal testbed to support such advanced research directions.

\section{Related Work}

\textbf{Collaborative Problem Solving (CPS)} is a core educational competency in which learners build shared understanding, combine knowledge and skills, and work jointly toward a common goal \cite{pisa2018assessment}. In mathematics education, open-ended tasks are widely recognized as supporting critical thinking and deep learning \cite{vygotsky1978mind}, and prior work has explored how technological scaffolds can support social interaction and equitable participation \cite{rose2008analyzing}. Recent studies have also begun to use LLMs to support virtual collaborative environments, such as MathVC \cite{yue2024mathvc}.
However, these frameworks usually rely on the theoretical frameworks~\citep{pisa2018assessment}. In contrast, our work provides a corpus composed entirely of real human students and adopts a bottom-up inductive approach driven by LLMs to discover a dialogue act taxonomy that better reflects the authentic micro-processes of physical classrooms.

\paragraph{Datasets and LLM Applications for Education}
Prior work has long explored conversational agents for supporting collaborative learning. For example, \citet{kumar2010architecture} proposed an architecture
for building conversational agents that support multi-learner collaborative
learning settings. More recently, LLMs have been widely applied in education~\citep{kasneci2023chatgpt}. The community has introduced several important dialogue datasets in recent years. These include TalkMoves \cite{suresh2022talkmoves}, which focuses on analyzing classroom teacher instructional discourse; NCTE \citep{demszky-hill-2023-ncte}, an elementary-school math-classroom dialogue dataset collected by the National Center for Teacher Effectiveness (NCTE); CIMA \cite{stasaski2020cima}, designed to support tutoring dialogue generation; and MathDial \cite{macina2023mathdial}, which simulates human teacher-student interactions grounded in math word problems.
However, they focus on the interaction paradigm of ``Teacher-Student''. 
Currently, there remains a severe lack of corpora and modeling efforts specifically targeting authentic horizontal ``Peer-to-Peer'' collaborative processes, and \benchmark fills this gap. In addition, our work showcases the potential of leveraging LLMs as powerful qualitative ``analytical engines'', assisting educational researchers in automating discourse analysis and insight discovery.

\section{Conclusion}

We present \benchmark, a new dataset of peer CPS dialogues from authentic middle-school mathematics learning environments, together with a corpus-grounded dialogue act taxonomy for analyzing collaborative interaction. Our experiments show that the dataset supports fine-grained study of collaborative dynamics, links between student traits and observed behaviors, and evaluation of LLM-based student simulation. We hope \benchmark will serve as a useful resource for future NLP and education research on collaborative learning processes.

\section{Limitations}
First, \benchmark is collected from a single math-centered summer camp with a middle-school population, so the dataset does not fully represent the broader range of K-12 classrooms. Limited by the size and duration of the camp, the dataset we collected covers a limited number of students and dialogues. However, considering the lack of resources in peer interaction for math learning, our dataset still provides unique value to the community. Meanwhile, we encourage future studies to further examine our findings based on a larger population of peer interaction data.
Second, our analyses and the scope of \benchmark focus on spoken dialogue and do not fully capture other important aspects of collaborative problem solving, such as gesture, gaze, shared writing, and other multimodal forms of coordination. This was mainly due to the technical difficulty in capturing synchronized multimodal data (e.g., recording students' shared math plotting synchronized with their conversation).
Finally, while we seek to demonstrate the broad applications of \benchmark, each of the three applications is worth deeper exploration. For example, for LLM student simulation (Application III), future work should look into how the current simulation diverges from the dataset labels, whether this divergence is caused by the underspecification of student character or an inherent LLM bias, etc.

\section*{Acknowledgments}
This work is supported by the National Science Foundation under award no. NSF-2418580 and NSF-2418582. We thank the Institutional Review Board (IRB) at George Mason University and William \& Mary for the review and support of this research.

\bibliography{custom}

\newpage
\appendix

\section{Student Demographic Information}
\label{appendix: student demographic}
The summer camp was held during the summer of 2025, supervised by the university IRB. In total, 38 students participated in the program. The camp was organized into two classrooms: one classroom with 16 sixth-grade students (except 1 fifth-grade talented-class student) and one classroom with 12 seventh-grade students and 10 eighth-grade students. The grade level was decided based on where they would enter in the 2025--2026 school year.
The gender distribution included 26 female students (68\%) and 12 male students (32\%). Students' self-reported ethnic backgrounds included Asian or Asian American ($n=16$), Black or African American ($n=7$), White or European ($n=6$), Hispanic or Latino/a ($n=4$), Middle Eastern or North African ($n=2$), Asian Indian ($n=1$), White/European/Hispanic (multi-identified; $n=1$), and prefer not to say ($n=1$). Each classroom was facilitated by two teachers, all of whom had approximately 20 years of teaching experience on average. Student participants were also asked to fill out a survey form about their background and personality.

After data cleaning, we have 27 students who have both the survey information and the dialogue history. Therefore, we pick these 27 students' discussion in \benchmark.

The dialogue data were collected from seven collaborative mathematics tasks:
\textit{Baker's Cake}, \textit{Ice Cream}, \textit{Znorlian},
\textit{1--100}, \textit{1001 Pennies}, \textit{Tax Collector}, and
\textit{Traffic Jam}. These seven tasks correspond to 55 transcript sessions in total, including
11 sessions for \textit{1001 Pennies}, 9 for \textit{Baker's Cake}, 9 for
\textit{Znorlian}, 8 for \textit{Ice Cream}, 8 for \textit{Tax Collector},
5 for \textit{1--100}, and 5 for \textit{Traffic Jam}.

\section{Action Taxonomy}
\label{appendix: action taxonomy}
This appendix presents the full corpus-grounded CPS dialogue act taxonomy used for turn-level annotation. The taxonomy is organized into six broad stages. Here, the term \textit{stage} refers to a functional grouping of interactional moves rather than a fixed temporal sequence in collaboration. Each dialogue act is defined below, together with representative examples from the corpus.

\subsection{Stage 1. Affect and Off-Task Talk}

\textbf{Stage definition.}
Actions used for emotional reactions, playful or evaluative commentary, and clearly social or off-task remarks that recur in classroom discourse but do not directly advance the mathematical collaboration.

\paragraph{Stage 1 Action 1. Off-Task or Affective Commentary}
\textbf{Definition.}
Making a brief affective reaction, playful remark, joke, evaluative exclamation, or clearly off-task social comment that does not primarily advance the mathematical work or direct collaboration.

\textbf{Examples.}
\begin{itemize}
    \item ``Well, clearly, he has stolen other people. He's making me lose brain cells. He's subtracting our brain power.''
    \item ``Because... like two days ago, I went to Denmark. I mean, Delaware. Who Delaware.''
    \item ``Oh, you've got to be kidding me.''
    \item ``I thought we were going to be like coding and stuff, not math. This is basically new summer school.''
    \item ``It's me, hi. I know what the problem it's you.''
\end{itemize}

\subsection{Stage 2. Communicating and Coordinating}

\textbf{Stage definition.}
Actions used to explain and justify reasoning for others, present group work publicly, manage participation or work on the shared artifact, and handle lightweight interactional turns that support collaboration without adding substantive mathematical content.

\paragraph{Stage 2 Action 1. Acknowledging or Receiving}
\textbf{Definition.}
Briefly accepting, confirming, or showing receipt of a prior contribution without adding substantial new mathematical content, such as short backchannels or uptake moves like ``yeah,'' ``okay,'' ``oh,'' or ``uh-huh.''

\textbf{Examples.}
\begin{itemize}
    \item ``Uh, huh.''
    \item ``Yeah.''
    \item ``Okay.''
    \item ``Yup.''
    \item ``Oh, yeah.''
\end{itemize}

\paragraph{Stage 2 Action 2. Explaining or Justifying Reasoning}
\textbf{Definition.}
Giving the rationale that connects a value, operation, representation, or method to the problem structure so others can follow why it makes sense.

\textbf{Examples.}
\begin{itemize}
    \item ``1.5 plus 1.5 would equal... because it's half of the square, that's 1.5 and not 2. And that means the entire square for one-fourth would have to be equal to four, but that's not going to be four.''
    \item ``That is three. That's three because if it was split in half, this split in half. This would be six, but split in half again would be three.''
    \item ``So over here, there are a bunch of cents. So there are, in one group of three, six numbers, 41, 41, 41, 41. 41 times four, 164, and 37 times two, 74, add it together equals 238 cents in one group of three. Well, one group of three pairs of numbers.''
    \item ``It's still half like the other one. That's still half, and that's still like each one fourth. So, nothing changed, except like the way it was shown.''
    \item ``So, let's say we have 19 cents, so that's 19 cents, and then so we have 919 cents, so that's \$9.19. So if we have 9919 cents then we have \$99.19.''
\end{itemize}

\paragraph{Stage 2 Action 3. Guiding Peer Participation or Shared Workspace Actions}
\textbf{Definition.}
Directing who should contribute next or what should be drawn, written, compared, erased, or discussed on the shared workspace.

\textbf{Examples.}
\begin{itemize}
    \item ``Who's next? Um, you.''
    \item ``Okay, erase this and write plus that.''
    \item ``You? So, you're the scarecrow this time.''
    \item ``All right, that's one. Write that down.''
    \item ``Can you write 228, 228.''
\end{itemize}

\paragraph{Stage 2 Action 4. Presenting Group Reasoning to an External Listener}
\textbf{Definition.}
Summarizing the group's process and answer for a teacher, class, verifier, or other audience outside the immediate back-and-forth collaboration.

\textbf{Examples.}
\begin{itemize}
    \item ``Okay, so, first we wrote a couple pennies here, and then we realized, started using the information we had to, like, organize, like which one is which. For example, we started with penny, nickel, dime, quarter, and then we realized, the second one, every two would be nickel. So, we first did that, and we filled out all the nickels here, and then we did every three would be dime. Then we realized they overlapped with the nickels, then Mrs. P told us that you're going to have to replace them.''
    \item ``We followed like the keys. So we just listed it out, like a bunch of pennies, and then we swapped out all the coins for like the numbers.''
    \item ``I can go. So, and then we did, we did 1001 divided by 12, and we got 83, with five coins left.''
    \item ``No, we realized, Mrs. P told us, what five coins. We added the first five coins, penny, nickel, dime, quarter, plus a penny, should equal 42. So, then we would add 42 here. At first, I thought this was the answer, then I realized the remainder. So we added 42 as the remainder and got 99.19.''
    \item ``And then we added all the coins together, and we got 119 cents.''
\end{itemize}

\subsection{Stage 3. Exploring and Understanding}

\textbf{Stage definition.}
Actions used to establish shared understanding of the task, lesson focus, givens, constraints, and the meaning of peers' statements before or while the group commits to a mathematical path.

\paragraph{Stage 3 Action 1. Asking for Clarification or Explanation}
\textbf{Definition.}
Requesting clarification about the task setup, a representation, an operation, or a peer's reasoning so the group can align on understanding.

\textbf{Examples.}
\begin{itemize}
    \item ``Wait, how did we get two? How did we get two?''
    \item ``How do you know that?''
    \item ``Why are you putting a line?''
    \item ``And then we're putting it into, into. Are we dividing 1001 by that, or something?''
    \item ``Where'd you get the 14 plus two?''
\end{itemize}

\paragraph{Stage 3 Action 2. Identifying Relevant Elements and Constraints}
\textbf{Definition.}
Pointing out or restating key symbols, pieces, quantities, goals, rules, or structural givens so the group shares what must be tracked.

\textbf{Examples.}
\begin{itemize}
    \item ``First, let's write down all the little, um, all the symbols we see. So I think we should an up, an up arrow. Triangle. And then this weird, like cross thing, and then, like a colored in square.''
    \item ``Oh, yeah. So we have one, two, eight, and then we're missing four.''
    \item ``Yeah, cause the question says, what is the shortest way to write 21.''
    \item ``It's every third one.''
    \item ``We're not finding out these two groups specifically, we're finding out how much cents are in those 1001 coins.''
\end{itemize}

\paragraph{Stage 3 Action 3. Orienting to Lesson or Activity}
\textbf{Definition.}
Framing what kind of activity, topic, or phase is happening next, including teacher setup about what students are about to learn or do, rather than naming specific mathematical givens inside the problem.

\textbf{Examples.}
\begin{itemize}
    \item ``I feel like, for all these tasks, like, we have to, like, really understand the problem, and then we can solve it.''
    \item ``The up is done. Adding numbers in Znorlian, and... is an addition sentence adding, what?''
    \item ``So we have the exact same riddle. Easiest way to find 21.''
    \item ``Okay. She said to do something. So I don't know. Do you know that?''
    \item ``Okay... now we have to find the shortest way to walk, write 21.''
\end{itemize}

\subsection{Stage 4. Monitoring and Revising}

\textbf{Stage definition.}
Actions used to evaluate whether current reasoning fits the task, interpret intermediate quantities, locate errors, and adjust the approach or answer accordingly.

\paragraph{Stage 4 Action 1. Challenging or Diagnosing an Error}
\textbf{Definition.}
Pointing out that a peer's idea, representation, operation, or assumption is wrong and, when possible, identifying the specific source of the problem.

\textbf{Examples.}
\begin{itemize}
    \item ``Oh yeah, no cause it can't be. It can't be because this one would have to be. This, when you do it, this one's three, and half of three is not two, it's 1.5.''
    \item ``No, because 12, minus two doesn't equal four, that equals 10.''
    \item ``That is half of one fourth, not a third.''
    \item ``No, no, no, there's no line in the middle.''
    \item ``Well, how is one times 125 equal 185?''
\end{itemize}

\paragraph{Stage 4 Action 2. Checking What a Number or Step Represents}
\textbf{Definition.}
Asking for or providing the meaning of an intermediate number, unit, or step to ensure quantities are being interpreted consistently.

\textbf{Examples.}
\begin{itemize}
    \item ``Huh, what's 125? This equals 125? Does this equal 125?''
    \item ``Two dollars and 38 cents.''
    \item ``Two times two times two.''
    \item ``Seven, but what are the values?''
    \item ``The remainder of five coins.''
\end{itemize}

\paragraph{Stage 4 Action 3. Checking Whether a Result Fits Constraints}
\textbf{Definition.}
Testing whether a proposed pattern, value, or answer is consistent with totals, fractions, repeating structure, or other task constraints.

\textbf{Examples.}
\begin{itemize}
    \item ``Okay so, this is um, um, um, what the heck, 1.5. Okay, so that's 3.75. Okay, so now we add the 2.25, which should equal six. Right? That is correct. So your thing is correct.''
    \item ``No, cause it wouldn't add up to \$12.''
    \item ``That's three. That's three. And then invisible line. Okay, that's not even even. Nuh uh.''
    \item ``So, we can't have a half a coin?''
    \item ``Because 5.5 would be ten... No, eleven. Plus three would be 14.''
\end{itemize}

\paragraph{Stage 4 Action 4. Revising the Method or Answer}
\textbf{Definition.}
Changing the current answer, method, or shared written work after a contradiction, objection, or diagnosed error has been recognized.

\textbf{Examples.}
\begin{itemize}
    \item ``I'm gunna change my answer, three. It's three. It's been three the whole time.''
    \item ``No, you're not supposed to write this. You're writing 16 plus four plus one.''
    \item ``Uh, one third. No, one fourth.''
    \item ``I'm changing the numbers because the math's wrong.''
    \item ``Yeah, there's three different. Oh wait, yeah. There's three different things, so I have to count for three? How many, we have to figure out groups of six, not groups of four.''
\end{itemize}

\subsection{Stage 5. Planning and Executing}

\textbf{Stage definition.}
Actions used to choose solution methods, advance candidate answers, strategically reorganize the problem, and carry out computations.

\paragraph{Stage 5 Action 1. Carrying Out a Computation Step}
\textbf{Definition.}
Performing or directing arithmetic or symbolic operations needed to execute the chosen strategy, including summing pattern units, dividing to find repeats, or combining known values.

\textbf{Examples.}
\begin{itemize}
    \item ``12 divided by four is three.''
    \item ``So, two plus eight is ten, plus four is 14. So, something minus two is 14.''
    \item ``1.5 plus 0.75, 2.25. That's 2.25. That one's 0.75.''
    \item ``Eight times four is 32.''
    \item ``No, this is two plus eight, which is ten, plus four is 14.''
\end{itemize}

\paragraph{Stage 5 Action 2. Decomposing or Repartitioning the Structure}
\textbf{Definition.}
Redrawing, splitting, or grouping the representation into easier subparts to make inference or calculation more manageable.

\textbf{Examples.}
\begin{itemize}
    \item ``And then, here, you just put a line through. Just put a line through. That's half of that. So, that will be 1.5 right?''
    \item ``Okay, so then this, this would be three and three, but that can't. So draw a line right here and right here. No right here, right here, just from here. No, no, no, here to here. Okay, so these are meant to be all equal squares. So these are all equal, and hold on can I be scarecrow for a sec? This... is three. This is all... three. So that is a... fourth, a fourth. What's a fourth of three? I think that's right, I don't know. What's a fourth of three? Go. A fourth of three.''
    \item ``Okay, how do we split this exactly? Yeah, you should put it right here. There and... there's another line, and just go right there. There. So, we know this is all six, and this all six. So, the easy part is, so this, it has to be... 0.75. So, 0.75 in here, and put imaginary lines over here. Wait. Do the uh... yeah. No, here first and then here. So this is 0.75, that's 0.75, that's 0.75. What are you doing exactly? Never mind. Okay, so that means this. This one, three. If we do that here too, it means 1.5. So, this is all 0.75, which would be three. This is three right here. So, put a three over here. So, this is a total of three. Oh, sorry. And then, this and three. This is all three minus 0.75, which would be um... 2.75. No, 2.25. So, this is 2.25.''
    \item ``The bottom one is six. Okay, I know something. Okay, if we have. Little lines right here. It would be split in half. So it would be 1.5 because 1.5 plus 1.5 is three, and this would be like a quarter so it would be three. I'm guessing this is. So this would be. Is six. Basically.''
    \item ``Okay so, okay so, put that in half. So, that we know is six. That's six. Six!''
\end{itemize}

\paragraph{Stage 5 Action 3. Proposing a Candidate Solution}
\textbf{Definition.}
Putting forward a tentative answer, value, decomposition, or completed expression for the group to consider.

\textbf{Examples.}
\begin{itemize}
    \item ``I think it should be \$4.''
    \item ``16 plus five.''
    \item ``Eight times three minus three.''
    \item ``\$3 per slice.''
    \item ``Oh, wait. I think I know it. Three. Three and 4.5. 4.5.''
\end{itemize}

\paragraph{Stage 5 Action 4. Proposing a Solution Strategy}
\textbf{Definition.}
Suggesting a method or plan for how to solve the problem, including how to decompose it, scale from a pattern, or proceed through the next mathematical steps.

\textbf{Examples.}
\begin{itemize}
    \item ``Yeah, we might be able to find the pattern, and then multiply.''
    \item ``Okay, first things first, we'll write what we know. Like, for example...''
    \item ``Yeah, that's where the pattern ends. The pattern ends at the 12 mark. If we add this, then we have to find how many times 12 goes into 1001. So we have to add this until like there, and we have to find out how many times that goes into 1001.''
    \item ``What if we just like, what if we just like. So, we have 1001, right? Subtract how much amount of pennies will take over that. Every second one will be a one.''
    \item ``So you have to do 250, divided by four, since 41, 41 appears every four different sections of the dollar.''
\end{itemize}

\subsection{Stage 6. Representing and Formulating}

\textbf{Stage definition.}
Actions used to construct, interpret, and refine shared representations, patterns, and structural models that make the problem mathematically workable.

\paragraph{Stage 6 Action 1. Constructing or Refining a Representation}
\textbf{Definition.}
Creating, extending, or modifying a shared drawing, list, grouping, sample case, or auxiliary visual structure to make hidden relationships visible.

\textbf{Examples.}
\begin{itemize}
    \item ``And then, here, you just put a line through. Just put a line through. That's half of that. So, that will be 1.5 right?''
    \item ``Wait, I already, I already know it. So, square. Wait so, square right here, square right here. There's one right here, tall box right here, rectangular box right here, and there's... oh wait. And then there's. So then, there's a line. Wait no, that's not even. The line should be like right here. Three over here and three over here. So then, this is all, if this is all three then that means this is 1.5. How would it be two? So then, one of them would be 0.75. So, 0.75, 0.75. That means this also has to be 0.75. Plus this, which is right here. Wait, this line is supposed to be up here though. Let's just draw this again. Half like this? Oh. And then, there's a line right here. And then, there's two boxes right here and a box right here. There! Wait, no. Yeah. So, this is 0.75, 0.75, don't judge my fives. What did I just say. What did I just say. Trying my best here.''
    \item ``Or maybe, to help you, we just put one, two, three, four, five, six.''
    \item ``Okay. One, two, three, four, five, six, seven, eight, nine, ten. Okay, um, P, N, P, N, P, N, N, N. Okay, wait, okay, yeah, yeah. Okay, um...''
    \item ``See, like that. Get it? Okay, and then quarters, every fourth one.''
\end{itemize}

\paragraph{Stage 6 Action 2. Mapping Representation to Values}
\textbf{Definition.}
Assigning numerical or relational meaning to symbols, shapes, pieces, or units based on givens and discovered relationships.

\textbf{Examples.}
\begin{itemize}
    \item ``One cent. Star is a nickel, five cents. Triangle is a dime, 10 cents. Heart is a quarter, 25.''
    \item ``This symbol is a four. Um, this symbol is a one... uh, this symbol is an eight.''
    \item ``One eighth. This is one eighth of the cake.''
    \item ``A half.''
    \item ``That means these mean one.''
\end{itemize}

\paragraph{Stage 6 Action 3. Proposing a Pattern or Structural Model}
\textbf{Definition.}
Suggesting a candidate pattern, grouping, equivalence, repeat unit, or structural interpretation that could organize the problem and guide solution work.

\textbf{Examples.}
\begin{itemize}
    \item ``So, the pattern is it doubles.''
    \item ``This grouping four... and one, two, three, four, these four. Okay, then, then I see two different groups. One with these and one with these.''
    \item ``That one's four, that's a 25. I think that's where the pattern repeats.''
    \item ``Oh. Two of these equals this, which equals that.''
    \item ``So there are, basically, two different ones. There's 41 and 37, and 41 and 41.''
\end{itemize}

\paragraph{Stage 6 Action 4. Reusing a Previous Case or Pattern}
\textbf{Definition.}
Using an earlier solved example, repeated structure, or analogous case to interpret the current representation or infer the current one.

\textbf{Examples.}
\begin{itemize}
    \item ``I think it is. It's the exact same thing, cause it's in half. Three, three, six. It should be the same price cause there's three slices, and then it's doubled again.''
    \item ``Oh. So, this is the exact same thing we get like with different shapes, but it's still in half and... in half.''
    \item ``Start drawing the square, right now. Oh wait, this is. Hold on, we recognize that right? That is half and half. Okay, so that is... half of six, three. So, three.''
    \item ``It's the same thing. That's also 1.5. That's 1.5. That's six. That's 4.5. It's the same thing. It's the same thing.''
    \item ``Same thing. Same as day three.''
\end{itemize}

\section{Expert Review of Action Taxonomy and Turn-level Assignment}
\label{appendix: expert review}

To evaluate the quality of the induced taxonomy and its turn-level labeling, we conducted an expert review. This evaluation focused on two questions. First, whether the dialogue act taxonomy induced by the LLM adequately captures the major dialogue actions that occur in students' ordinary discussions; for this purpose, we provided the complete taxonomy definitions together with multiple examples. Second, whether the turn-level labels assigned by the LLM based on the taxonomy were accurate, and, if not, whether the problems arose because the taxonomy had failed to include a necessary action type or because the appropriate action type existed but had been assigned incorrectly.
Based on this goal, we constructed an evaluation set containing 100 instances of conversational turns. Among them, 50 instances were randomly sampled from the full corpus to assess labeling quality under typical conditions, and the other 50 instances were drawn from the turns that the LLM marked as low-confidence in the second round of labeling. This design allows the evaluation to cover both ordinary cases and the subset of examples that remain most uncertain under the final taxonomy and are therefore most likely to reveal residual labeling difficulties.

Two doctoral students who majored in Math Education and had multiple years of school teaching served as experts. During the review, each instance was presented to experts in a standardized format, including the speaker identity, the necessary dialogue context, the current turn to be judged, the action assigned by the LLM, and the model's brief rationale. The experts were asked to judge whether the assigned act adequately described the student's turn. If not, they were further asked whether a better label could be found among the existing candidate actions; if none of the existing acts fit, they were allowed to propose a new action label. In this way, the evaluation jointly examined both assignment quality and taxonomy coverage.

The results indicate that the taxonomy provides strong coverage of the major interactional behaviors in the corpus. The experts also endorsed the definitions of the extracted dialogue act taxonomy. Among the 50 randomly sampled instances, experts judged all assigned dialogue acts to be appropriate. Among the 50 low-confidence instances, 11 were judged to be assigned incorrectly. Importantly, in all of these error cases, the problem was identified as an assignment error rather than a missing act type in the taxonomy; that is, experts were always able to identify a relatively more suitable label from the existing candidate actions, without introducing any new category. This result suggests that the main remaining source of error lies in turn-level label assignment rather than insufficient coverage of the action inventory itself. Overall, the expert review supports the validity of the induced taxonomy and suggests that this LLM-assisted pipeline is able to capture the major CPS interactional behaviors in the corpus with substantial coverage.

\section{Survey-based Student Profiles}
\label{appendix :survey_profiles}

Student profile groupings used in Application II are derived from a self-report survey administered in the beginning of the camp program. The survey contains 20 items, each answered on a five-point Likert scale: \textit{strongly disagree}, \textit{disagree}, \textit{neutral}, \textit{agree}, and \textit{strongly agree}. The items cover multiple aspects of students' orientations toward mathematics learning and peer collaboration, including math confidence and enjoyment, collaborative preference, leadership tendency, curiosity, organization, sociability, conflict avoidance, and concern about making mistakes. For the analyses in the main text, we organize related items into broader profile dimensions such as \textit{confidence}, \textit{collaboration}, and \textit{leadership}. For each dimension, students are divided into two broad groups based on whether their responses indicate overall endorsement of that dimension (i.e., more agreement-oriented responses) or not. These groupings are intended as interpretable survey-based typologies for descriptive analysis, rather than as formal psychometric scales.

\paragraph{Survey items.}
The survey includes the following items:
\begin{itemize}
    \item I enjoy learning math.
    \item I feel confident solving math problems.
    \item I often get frustrated when working on math.
    \item I like explaining math ideas to others.
    \item I find math to be useful in my daily life.
    \item I prefer working alone on math tasks.
    \item I feel comfortable asking questions in a group setting.
    \item Having others' perspectives often helps me learn better.
    \item Working in a team makes learning more engaging.
    \item I like being the center during the group discussion.
    \item I often listen to others before sharing my ideas.
    \item When I have an idea that others disagree with, I will persuade them.
    \item I often help assign tasks and resolve conflict in a team.
    \item I learn new things quickly.
    \item I'm a curious person and I like trying new things.
    \item I keep track of what needs to be done carefully.
    \item I like interacting with a lot of people.
    \item I try to avoid arguments, even if I disagree.
    \item I often worry about making mistakes in front of others.
\end{itemize}

\end{document}